\documentclass[lettersize,journal]{IEEEtran}
\usepackage{amsmath,amsfonts}
\usepackage{algorithmic}
\usepackage{algorithm}
\usepackage{array}
\usepackage[caption=false,font=normalsize,labelfont=sf,textfont=sf]{subfig}
\usepackage{textcomp}
\usepackage{stfloats}
\usepackage{url}
\usepackage{verbatim}
\usepackage{graphicx}
\usepackage{cite}
\usepackage{kotex-logo}
\usepackage{wrapfig}
\usepackage{graphicx}
\usepackage[table,xcdraw]{xcolor}
\usepackage{xcolor}
\usepackage{booktabs}
\usepackage{newfloat}
\usepackage{listings}
\usepackage{multirow}
\usepackage{multicol}
\usepackage{lipsum}
\usepackage{standalone}
\newcommand{\Tref}[1]{Table~\ref{#1}}

\newcommand{\Fref}[1]{Figure~\ref{#1}}

\usepackage{kotex}
\usepackage{soul}

\begin{document}

\title{Bridging the Domain Gap: A Simple Domain Matching Method for Reference-based Image Super-Resolution in Remote Sensing}

\author{Jeongho Min,~Yejun Lee,~Dongyoung Kim,~Jaejun Yoo

\thanks{Jeongho Min, Yejun Lee, Dongyoung Kim and Jaejun Yoo\textsuperscript{\dag} are with the Graduate School of Artificial Intelligence, Ulsan National Institute of Science and Technology, Ulsan, 44919, Republic of Korea e-mail: \{jeongho.min, yejun.lee, dykim94, jaejun.yoo\}@unist.ac.kr}

\thanks{$\dagger$ : Corresponding author}}



\maketitle
\begin{abstract}
Recently, reference-based image super-resolution (RefSR) has shown excellent performance in image super-resolution (SR) tasks. The main idea of RefSR is to utilize additional information from the reference (Ref) image to recover the high-frequency components in low-resolution (LR) images. By transferring relevant textures through feature matching, RefSR models outperform existing single image super-resolution (SISR) models. 
However, their performance significantly declines when a domain gap between Ref and LR images exists, which often occurs in real-world scenarios, such as satellite imaging. In this letter, we introduce a Domain Matching (DM) module that can be seamlessly integrated with existing RefSR models to enhance their performance in a plug-and-play manner. To the best of our knowledge, we are the first to explore Domain Matching-based RefSR in remote sensing image processing. Our analysis reveals that their domain gaps often occur in different satellites, and our model effectively addresses these challenges, whereas existing models struggle. Our experiments demonstrate that the proposed DM module improves SR performance both qualitatively and quantitatively for remote sensing super-resolution tasks. 
\end{abstract}

\begin{IEEEkeywords}
Reference-based Image Super Resolution, Domain Adaptation, Remote Sensing
\end{IEEEkeywords}

\section{Introduction}
\IEEEPARstart{I}{mproving} the resolution of satellite imagery can enhance the performance of sub-tasks such as  {semantic segmentation}~\cite{diakogiannis2020resunet}, object detection~\cite{han2014object}, change detection~\cite{chen2021remote}, and data fusion~\cite{gao2023cross}. It can also find active utilization in military applications or the assessment of damage extent resulting from natural disasters. However,
obtaining high-resolution satellite data inherently comes with significant costs, particularly in terms of hardware investments. As a result, the software-based improvements achieved by applying super-resolution techniques on satellite data hold immense potential for generating substantial economic value. 

Single image super-resolution (SISR) methods that reconstruct high-resolution (HR) images from a single low-resolution (LR) image have demonstrated excellent performance in remote sensing tasks
~\cite{pan2019super,zhang2020scene,zhang2020remote,wang2023remote}. 
However, SISR algorithms face challenges when applied to images outside of the training data, since these algorithms are often trained on specific types of image distributions. To address these challenges, numerous works ~\cite{kang2022multilayer, luo2022learning} have been proposed to enhance the efficacy of SISR methodologies. However, SISR models still have a fundamental limitation in that they only rely on a single LR image to estimate the HR image.

Unlike SISR models, reference-based super-resolution (RefSR) models incorporate an additional high-resolution reference (Ref) image. A typical RefSR framework consists of two major parts:  correspondence matching and texture transfer. Correspondence matching involves finding the most similar correspondence between the input and Ref images, while texture transfer entails transferring features from the reference image to the input feature based on their similarity. By exploiting  the HR Ref images that are rich in texture and high-frequency information, RefSR models generally have shown superior performance than the SISR models \cite{zheng2018crossnet,zhang2019image,shim2020robust,jiang2021robust,lu2021masa,cao2022reference,dong2021rrsgan} on the benchmark dataset such as CUFED5 dataset~\cite{zhang2019image}.


However, the CUFED5 dataset is constructed by pairing closely matched images together, where the input and reference images can be considered as aligned paired data. It is important to note that such closely matched data scenarios are not commonly encountered in real-world situations. Moreover, this real-world data, acquired from different devices, presents additional challenges for the RefSR task compared to the idealized scenarios. We aim to perform super-resolution tasks using a real-world dataset, specifically a remote sensing dataset. One advantage of the remote sensing dataset is that we have access to paired images captured by different satellites on the same geographical region~\cite{dong2021rrsgan}. This allows us to utilize the reference images from the satellites to enhance the resolution of the input images through super-resolution techniques. However, there exists a large domain gap caused by variations in shooting angles, capture times, and other factors due to acquisition from different satellites. Consequently, there is no assurance that RefSR models will accurately match and effectively transfer genuinely informative textures in such situations. 

To address this challenge,  we propose a simple but effective method to reduce the domain gap between LR and Ref images. 
%
Specifically, we propose to perform a grayscale transformation on the input and Ref images to enhance the matching performance while preserving structural information. We then employ a style transfer-based approach to generate a reference image that is adapted to the distribution of the input image, enabling more effective texture transfer. 
Moreover, we leverage the renowned Whitening and Coloring Transform (WCT), a well-established method in the field of style transfer, to facilitate seamless texture transfer. However, since WCT is developed for artistic style transfer, it is not designed to preserve structural information. To avoid such distortion, we incorporate the Phase Replacement (PR) technique into the process, ensuring the preservation of structural information, which is crucial for our SR task.

As demonstrated by both quantitative and qualitative evaluations, our method brings remarkable enhancements over existing state-of-the-art (SOTA) RefSR models. Our results highlight the efficacy of our proposed method in mitigating the domain gap problem and improving the overall performance of RefSR in real-world scenarios, specifically for electro-optical imagery. Our contributions are summarized as follows:
\begin{itemize}
	
    \item  We provide a simple but effective method that can super-resolve an image well even when the Ref image exhibits a  different distribution from the input image.
    
    \item Our method has significantly improved existing models with 0.03 to 0.16 in PSNR for SR performance and 0.1 to 0.15 in SSIM-S for matching performance.
    
    \item Our method requires no extra training, making it easy to integrate into existing RefSR models.
\end{itemize}
\begin{figure*}[t]
\centering
\includegraphics[width=1.00\linewidth]{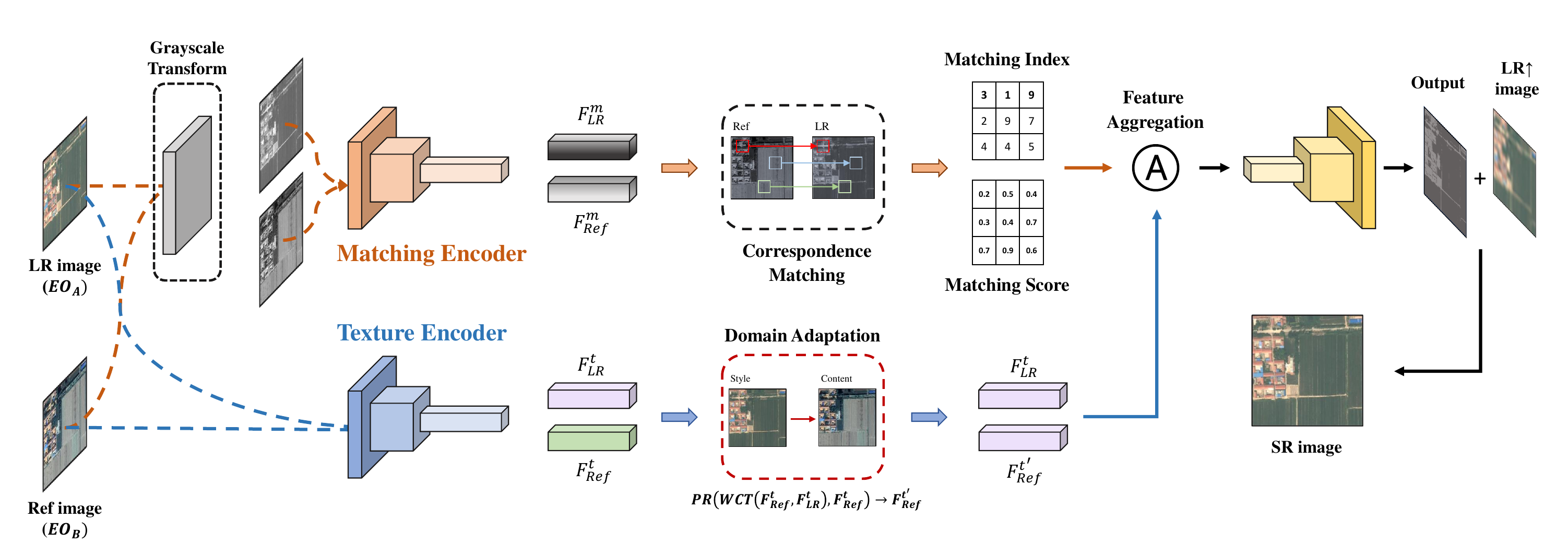}
\vspace{- 0.5 cm}
\caption{{\bf Overview} We applied a  grayscale transform to the input images before feeding it into the matching encoder during the correspondence matching process.  {In the texture transfer process, $F^t_{Ref}$ is the domain adapted by $F^t_{LR}$ to enhance the information of transferring textures. In this experiment, the domain adaptation module indicates WCT with phase replacement technique.} $EO_A$ and $EO_B$ refer to electro-optical images captured by different satellites. }
\label{fig:oveview}
\end{figure*}

\section{Analysis}
\noindent
In this section, we undertake a visual examination to assess the effectiveness of the correspondence matching process in the existing RefSR models and investigate the impact of the DM module on the matching process. 
Surprisingly, contrary to common belief, our findings reveal that the existing RefSR models demonstrate poor matching performance when applied to remote sensing datasets, which have a large domain gap. 


\smallskip
\noindent\textbf{Investigation of the matching quality.}
In RefSR, the correspondence matching process plays a crucial role in determining its final super-resolution (SR) performance. However, to the best of our knowledge, no previous study has conducted a comprehensive analysis of the matching results.
%
Previous studies~\cite{zhang2019image,jiang2021robust,lu2021masa,cao2022reference} have typically evaluated the impact of the correspondence matching process indirectly, through simple ablation studies tied to the final SR performance. However, improvements in the final SR performance may not necessarily stem from a superior matching module; other factors such as increased model capacity could also contribute. 

\begin{figure*}[t]
\begin{center}
\centering
\includegraphics[width=2.0\columnwidth]{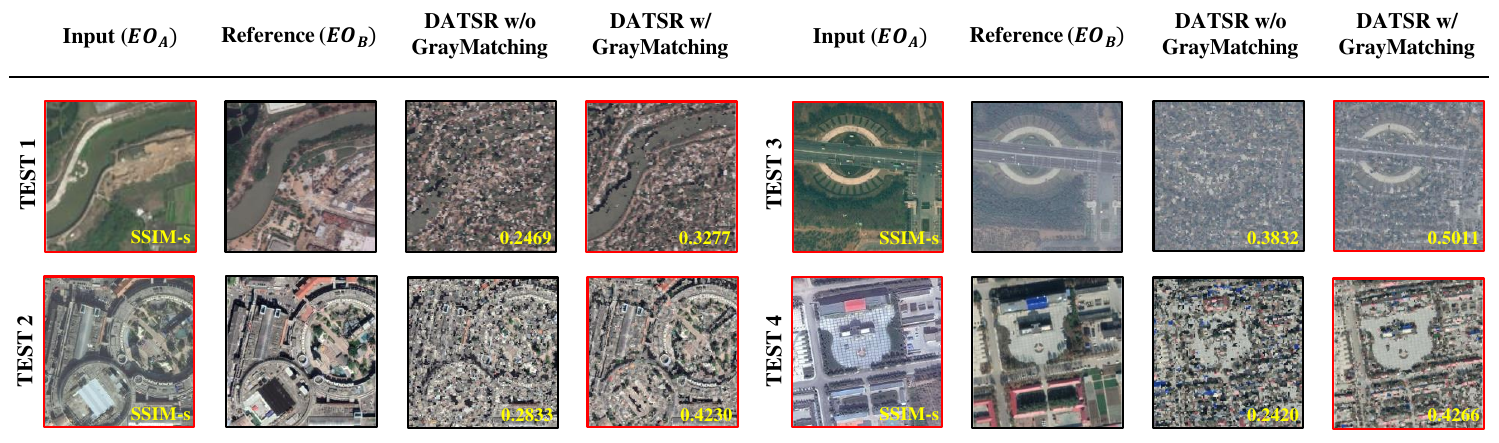}
\vspace{-0.5 cm}
\end{center}
   \caption{\textbf{ {Visualization of matching results of DATSR with and without grayscale transformation in the corresponding matching process.}} The matching result indicates the outcome after identifying the most relevant reference feature to the input, implying that a better matching is achieved as it becomes closer to the input. Our approach demonstrates a much cleaner and more effective retrieval of detailed information compared to the existing baseline. }   
\label{fig:match}
\end{figure*}

To directly assess the quality of matching and its influence on the final SR performance, we propose to undertake a detailed examination of the matching process and visualize the results for clarity (\Fref{fig:match}). The correspondence matching process is based on cosine similarity, which involves extracting the index and score of the Ref image that corresponds to each pixel in the input image. If the matched feature from the Ref image gives useful information, SR performance would ideally increase. 
In the case of EO data, the acquired data originates from the same location. Assuming minimal terrain variations and an ideally performed matching process, it is expected that patches from the same location would be successfully matched. Consequently, the matching result should ideally reflect the structural characteristics of the LR image. Therefore, evaluating the structural similarity between the matching result and the LR image becomes a means to assess the performance of correspondence matching and determine how well it preserves the structural properties. 

To evaluate the quality of the correspondence matching, 
we utilize the structure component $s$ of $\text{SSIM}(x, y) = l(x, y) c(x, y) s(x, y)$, which we denote as SSIM-s: 
\begin{align}
    \text{SSIM-s} : s(x, y)=\frac{\sigma_{xy}+C_3}{\sigma_x\sigma_y+C_3}, ~~C_3=\frac{(0.03L)^2}{2}
\end{align}
where $x$ and $y$ indicates LR image and matching result respectively, $l$ is luminance, $c$ is contrast, $\sigma_x^2$ and $\sigma_y^2$ are the variances of $x$ and $y$, and $\sigma_{xy}$ is the covariance between $x$ and $y$. Additionally, $L$ denotes the data range. A higher $s(x, y)$ value indicates that the two images have more similar structures or spatial dependencies. To the best of our knowledge, we are the first to explicitly evaluate the actual matching quality. 

\smallskip
\noindent\textbf{Visual interpretation of the matching procedure.} 
The result of the matching process  in \Fref{fig:match} is generated by reorganizing patches from the reference image based on these indices.  {Contrary to the widely held notion that correspondence matching modules should locate similar information, the structure of the matched results is fragmented, while clean matching results are obtained when grayscale transformation is used} (\Fref{fig:match}). We hypothesize that this result may be due to the visually disparate distributions (e.g., color, brightness, contrast) introduced by different acquisition devices, despite them capturing the same scene.

To validate our hypothesis, we develop a simple adjustment, a grayscale transformation, aimed at reducing basic visual disparity.  As shown in \Fref{fig:match}, the application of this grayscale transformation, which we term as Gray Matching, yields visually less noisy results compared to the baseline and more accurately reflects the structure information inherent in the LR image. This outcome indicates the performance improvement in the matching process achieved by implementing this simple adjustment. Later, we show that this enhancement in correspondence matching indeed translates into the performance enhancement of the final SR outcome (\Tref{table:ablation}). 

\section{Method}
\label{sec:method}
\noindent Inspired by our prior analysis, we propose Domain Matching (DM) modules that aim to minimize the domain gap between the source and reference images, while preserving crucial structural information: Gray Matching, Whitening and Coloring Transform (WCT), and Phase Replacement (PR).

\smallskip
\noindent\textbf{Gray Matching.} To reduce the domain gap in the correspondence matching step, we propose to introduce grayscale transformation for input images.
 {The grayscale transformation involves averaging three channels to create a single-channel image before it is fed into the encoder as input, prior to the Matching encoder process.} Applying grayscale transformation to an image effectively eliminates color information, leaving the structural information of the image. By this simple correction, we find that the impact of the distribution gap (e.g., hue, brightness) between two images is reduced effectively, and the matching performance is improved. 

\smallskip
\noindent\textbf{Whitening and Coloring Transform (WCT).}
To reduce the domain gap in the texture transfer step, we use Whitening and Coloring Transform (WCT)~\cite{li2017universal}. WCT matches the covariance matrix of the content feature map to the style feature map from the pre-trained VGG network: 
\begin{equation}
f_c \leftarrow f_c-m_c~,~~\hat{f}_c = E_cD_c^{-\frac{1}{2}}E_c^\top f_c
\end{equation}
\begin{equation}
\hat{f_{cs}}=E_sD_s^{\frac{1}{2}}E_s^\top \hat{f_c}~,~~\hat{f}_{cs} \leftarrow \hat{f}_{cs}+m_s
\end{equation}
$f_c$ and $f_s$ are feature maps of content images and style images from a certain layer of the VGG network. 
\textbf{1) Whitening} :
Whitening Transform is a linear transformation to make the covariance matrix of a feature map to the identity matrix. $D$ is a diagonal matrix for scaling, and $E$ is the corresponding orthogonal matrix of eigenvectors for rotation. $f_c$ is centered data by subtracting the sample mean $m_c$.
\textbf{2) Coloring} :
Coloring Transform is a reverse of a whitening process. It transforms the whitened data to have a specific covariance matrix. $E_s$ and $D_s$ are an orthogonal eigenvector matrix and diagonal matrix of $f_sf_s\top$. Then, by adding the mean of a style feature $m_s$ to $f_{cs}$, the output gets stylized of $f_s$.  


\smallskip
\noindent\textbf{Stylized feature map modification via Fourier perspective.}
Directly applying style transfer techniques may damage the structural information of the HR Ref image. To address this issue, we propose the Phase Replacement (PR) technique. Following \cite{jin2022style}, we preserve the phase information of the content feature map while keeping the amplitude of the stylized feature map during the WCT process. 
This approach allows us to retain the structural information of the image~\cite{jin2022style}:
\begin{equation}
{\cal F}^{cs}_{u,v}=|{\cal F}|^{cs}_{u,v}\odot \cos \angle{\cal F}^{c}_{u,v}+j|{\cal F}|^{cs}_{u,v}\odot\sin\angle{\cal F}^c_{u,v}.
\end{equation}
Here, $\odot$ denotes element-wise multiplication, and ${\cal F}^{cs}$ and ${\cal F}^{c}$ denote discrete Fourier transformed stylized feature map and content feature map, respectively. 
%
By utilizing this technique, we can effectively narrow down the domain gap between input and reference features while preserving the structural information, which eventually allows the RefSR model to achieve an improved SR performance.

\begin{figure*}[t]
\centering
\includegraphics[width=1.02\linewidth]{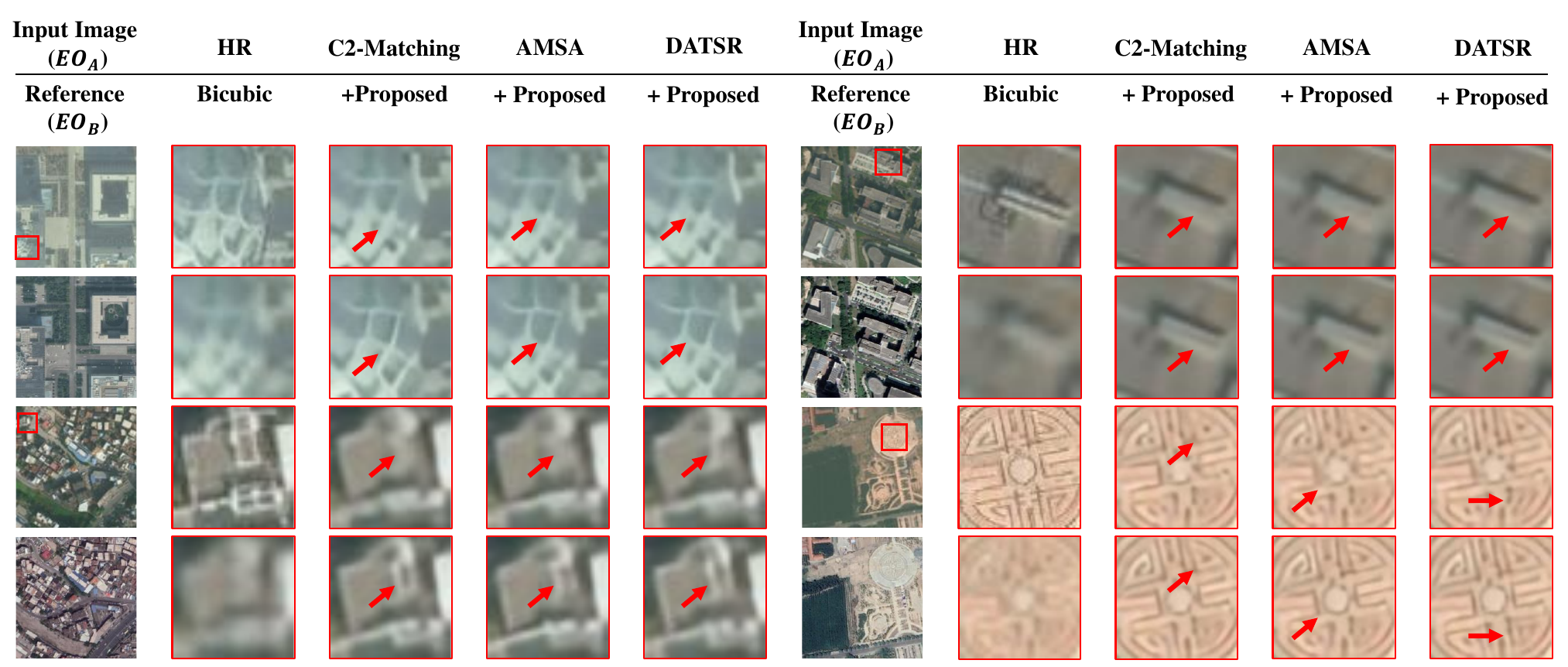}
\caption{Qualitative comparison of baseline RefSR models and proposed methods. Our proposed method shows better performance compared to previous SOTA models and recovers more details and shows good visual quality.}
\label{fig:Qualitative_rrs}
\end{figure*}

\section{Experiments}
\subsection{Experimental settings}
\noindent\textbf{Baseline models.} We use C2-Matching, AMSA, and DATSR as our baseline.  {We examine the SR performance on $\times4$ scale when adding our domain matching modules into these models. For training specifics and hyperparameter configurations, we follow the setting of C2-Matching}~\cite{jiang2021robust}. 

\noindent\textbf{Evaluation Metrics.} We measured the performance of the models using PSNR and SSIM, specifically in the Y channel of the YCbCr color space. Higher PSNR and SSIM are interpreted as better model performance. 


\noindent\textbf{Datasets.}
We utilize the RRSSRD dataset to train our RefSR models. The RRSSRD dataset comprises electro-optical data acquired from satellites and includes paired data captured from different satellites of the same scene. The training set consists of 4,047 paired 480×480 sized images. HR images in the dataset consist of Gaofen-2, WorldView, and Microsoft Virtual Earth 2018, while Ref images are sourced from Google Earth.

\subsection{Results on the dataset with domain gap}
\noindent We experiment with our baseline RefSR models (C2-Matching, AMSA, DATSR) under the existence of domain gaps. In an ideal scenario, the domain matching module should exhibit comparable performance to the baseline model in the absence of a domain gap, while demonstrating greater robustness than the baseline model in situations where a domain gap exists (\textit{i.e.}, exhibiting lower performance degradation). To verify this, we conduct experiments in remote sensing datasets. For the models, we trained them separately and evaluated them on their own test sets.

\smallskip
\noindent\textbf{Qualitative results.} 
In \Fref{fig:Qualitative_rrs}, we apply our proposed domain matching module to the three baseline models and compared their visual results. We test them on real-world images from the RRSSRD dataset, which has a significant domain gap between input and reference images. As we can see, the baseline models fail to reduce the domain gap, resulting in blurry outputs or the inability to successfully reconstruct textures. However, by plugging in our proposed domain matching module, we successfully reduce the domain gap and generate visually appealing results for all models.

\smallskip
\noindent\textbf{Quantitative Results.} \Tref{table:quan} presents the quantitative results, showing that our method significantly improves the performance of all state-of-the-art models on the RRSSRD dataset across all testing sets. Our method shows good generalization and easy applicability to different domains with a large domain gap, as evidenced by its strong performance on the 2nd and 4th testing sets. Furthermore, the comparison between C2-matching and DATSR, reveals that the latest SOTA model falls short of the RRSSRD dataset. This highlights the importance of addressing the domain gap in real-world datasets. \Tref{table:match_ssim} shows the comparison of SSIM-s when performing the corresponding matching with and without our method. This measures how well the reference matches the input in terms of similarity. This again demonstrates that improved matching performance translates into the performance enhancement of final SR outputs.

\begin{table}[t]
\caption{ {Quantitative comparison on RRSSRD Testset}\label{table:quan}}
\Huge
\centering
\resizebox{\columnwidth}{!}{
\begin{tabular}{ l c c c c c c c c c c } 
\toprule 
\multirow{2}{*}{Methods} & \multirow{2}{*}{Params} & \multicolumn{2}{c}{1st test set} & \multicolumn{2}{c}{2nd test set} & \multicolumn{2}{c}{3rd test set} & \multicolumn{2}{c}{4th test set} \\ 
\cmidrule(lr){3-4} \cmidrule(lr){5-6} \cmidrule(lr){7-8} \cmidrule(lr){9-10}
 & & PSNR$\uparrow$ & SSIM$\uparrow$ & PSNR$\uparrow$ & SSIM$\uparrow$ & PSNR$\uparrow$ & SSIM$\uparrow$ & PSNR$\uparrow$ & SSIM$\uparrow$ \\ 
\cmidrule(lr){1-10} 
C2-Matching & \multirow{2}{*}{8.9M} & 34.05 & 0.891 & 33.13 & 0.860 & 31.19 & 0.827 & 33.37 & 0.847 \\
+ Proposed  &   & {\color[HTML]{009901} \textbf{+0.03}} & {\color[HTML]{009901} \textbf{+0.000}} & {\color[HTML]{009901} \textbf{+0.10}} & {\color[HTML]{009901} \textbf{+0.001}} & {\color[HTML]{009901} \textbf{+0.14}} & {\color[HTML]{009901} \textbf{+0.003}} & {\color[HTML]{009901} \textbf{+0.12}} & {\color[HTML]{009901} \textbf{+0.001}} \\ 
\cmidrule(lr){1-10}
AMSA & \multirow{2}{*}{9.7M} & 34.12 & 0.891 & 33.16 & 0.860 & 31.24 & 0.828 & 33.43 & 0.847 \\ 
+ Proposed &    & {\color[HTML]{009901} \textbf{+0.04}} & {\color[HTML]{009901} \textbf{+0.001}} & {\color[HTML]{009901} \textbf{+0.05}} & {\color[HTML]{009901} \textbf{+0.001}} & {\color[HTML]{009901} \textbf{+0.07}} & {\color[HTML]{009901} \textbf{+0.002}} & {\color[HTML]{009901} \textbf{+0.07}} & {\color[HTML]{009901} \textbf{+0.002}} \\ 
\cmidrule(lr){1-10}
DATSR & \multirow{2}{*}{18.9M} & 33.98 & 0.890 & 33.03 & 0.858 & 31.13 & 0.826 & 32.28 & 0.845 \\
+ Proposed &    & {\color[HTML]{009901} \textbf{+0.12}} & {\color[HTML]{009901} \textbf{+0.001}} & {\color[HTML]{009901} \textbf{+0.16}} & {\color[HTML]{009901} \textbf{+0.003}} & {\color[HTML]{009901} \textbf{+0.18}} & {\color[HTML]{009901} \textbf{+0.003}} & {\color[HTML]{009901} \textbf{+0.16}} & {\color[HTML]{009901} \textbf{+0.002}} \\ 
\bottomrule
\end{tabular}
}
\end{table}

\begin{table}[t]
\caption{SSIM and SSIM-s comparison on RRSSRD}
\label{table:match_ssim}
\centering
\Huge
\resizebox{\columnwidth}{!}{%
\begin{tabular}{ l c c c c c }
         \toprule
         Methods & Metric & 1st test set & 2nd test set & 3rd test set & 4th test set \\\midrule 
         \multirow{2}{*}{DATSR} & \text{SSIM} $\uparrow$ &{0.2306} &{0.2372} &{0.2131} & 0.2195 \\  
         & \text{SSIM-s} $\uparrow$ &{0.3665} &{0.3771}  &{0.3347}  & {0.3170}  
          \\\midrule   
         \multirow{2}{*}{Ours} & \text{SSIM} $\uparrow$ 
         &\color[HTML]{009901}\textbf{+0.1256} &\color[HTML]{009901}\textbf{+0.1056} &\color[HTML]{009901}\textbf{+0.1523} &\color[HTML]{009901}\textbf{+0.1468} \\  
        & SSIM-s $\uparrow$ 
        &\color[HTML]{009901}\textbf{+0.1123} &\color[HTML]{009901}\textbf{+0.0981} &\color[HTML]{009901}\textbf{+0.1400} &\color[HTML]{009901}\textbf{+0.1374}
        \\\bottomrule
    \end{tabular}
}
\end{table}



\subsection{Ablation study}
\noindent In this section, we demonstrate the effectiveness of incorporating our module into the baseline model. As shown in \Tref{table:ablation} and \Fref{fig:match}, by utilizing Gray Matching in the corresponding matching stage, matching performance is enhanced compared to DATSR both qualitatively and quantitatively. 
Applying both methods including WCT and PR demonstrates a significant improvement in the results. Experimental results show that our method improves the performance of SOTA RefSR models.

\begin{table}[t]
\caption{ {Ablation study on our module on RRSSRD Testset}
\vspace{-3mm}
\label{table:ablation}}
\centering
\Huge
\resizebox{\columnwidth}{!}
{
\begin{tabular}{lccccc}                \\ \toprule Methods  & Metric & 1st test set & 2nd test set & 3rd test set & 4th test set \\ \cmidrule(lr){1-6}       
\multirow{2}{*}{\begin{tabular}[c]{@{}l@{}}DATSR\\ (Baseline)\end{tabular}} & PSNR$\uparrow$   
& 33.98      & 33.03      & 31.13      & 33.28      
\\ & SSIM$\uparrow$   
& 0.890      & 0.858      & 0.826      & 0.845      
\\  \cmidrule(lr){1-6}
\multirow{2}{*}{\begin{tabular}[c]{@{}l@{}}+ Gray \\ Matching\end{tabular}} & PSNR$\uparrow$   
& +0.08      & +0.1      & +0.15      & +0.13      
\\ & SSIM$\uparrow$   
& +0.000      & +0.002      & +0.003      & +0.002 
\\ \cmidrule(lr){1-6}
\multirow{2}{*}{+ $WCT_{PR}$}                          & PSNR$\uparrow$   
& +0.10      & +0.09      & +0.06      & +0.10      
\\ & SSIM$\uparrow$   
& +0.001      & +0.001      & +0.000      & +0.001      
\\ \cmidrule(lr){1-6}
\multirow{2}{*}{\begin{tabular}[c]{@{}l@{}}+ Gray Mat.\\ $AdaIN_{PR}$\end{tabular}}      & PSNR$\uparrow$ 
& +0.09 & +0.11 & +0.16 & +0.10 
\\ & SSIM$\uparrow$   
& +0.000      & +0.002      & +0.003      & +0.001 
\\ \cmidrule(lr){1-6}
\multirow{2}{*}{\begin{tabular}[c]{@{}l@{}} Ours
\end{tabular}} 
& PSNR$\uparrow$ 
& \color[HTML]{009901}\textbf{+0.12} & \color[HTML]{009901}\textbf{+0.16} & \color[HTML]{009901}\textbf{+0.18} & \color[HTML]{009901}\textbf{+0.16} 
\\ & SSIM$\uparrow$   
& \color[HTML]{009901}\textbf{+0.001}      & \color[HTML]{009901}\textbf{+0.003}      & \color[HTML]{009901}\textbf{+0.003}      & \color[HTML]{009901}\textbf{+0.002} 
\\ \bottomrule
\end{tabular}
}
\end{table}

\section{Conclusion}
\noindent We proposed a domain-matching module for the reference-based image super-resolution task, which can be implemented in a plug-and-play manner. We investigated the matching process of RefSR models and investigated the matching results. Based on our analysis of the matching process, we proposed the domain-matching processes in RefSR models. By aligning the domains in the correspondence matching and texture transfer, we showed that our method improves performance of each part when plugged into the existing RefSR framework.

\section*{Acknowledgements}
This work was supported by the Korea Research Institute for Defence Technology Planning and Advancement (KRIT) grant funded by the Korea government (DAPA) in 2022 (KRIT-CT-22-037, SAR Image Super-Resolution Technology for Improving of Target Identification Performance 50\%), National Research Foundation of Korea (NRF) grant funded by the Korea government (MSIT) (2022R1C1C100849612 20\%), Institute of Information \& communications Technology Planning \& Evaluation (IITP) grant funded by the Korea government (MSIT) (No.2020-0-01336, Artificial Intelligence Graduate School Program (UNIST), 5\%, No.2021-0-02068, Artificial Intelligence Innovation Hub, 5\%, No.2022-0-00959, (Part 2) Few-Shot Learning of Causal Inference in Vision and Language for Decision Making, 10\%, No.2022-0-00264, Comprehensive Video Understanding and Generation with Knowledge-based Deep Logic Neural Network, 10\%).

 

\bibliography{references}
\bibliographystyle{IEEEtran}

\end{document}